\documentclass[11pt,onecolumn,letterpaper]{article}

\usepackage[utf8]{inputenc}
\usepackage[T1]{fontenc}
\usepackage{times}

\usepackage{microtype}
\usepackage{graphicx}
\usepackage{booktabs}
\usepackage{tabularx}
\usepackage{amsmath,amssymb}
\newcommand{\rr}{\raggedright\arraybackslash\hyphenpenalty=10000\exhyphenpenalty=10000}
\newcolumntype{Y}[1]{>{\rr\hsize=#1\hsize}X}
\usepackage{xcolor}
\usepackage{hyperref}
\usepackage{cite}
\usepackage[margin=1in]{geometry}
\usepackage{caption}
\title{Damage-TriageFormer: A Foundation-Model Framework for Typology-Based Building Damage Assessment from Mono-Temporal Imagery}

\author{
  Yiming Xiao\textsuperscript{1,*} \qquad
  Yu-Hsuan Ho\textsuperscript{1} \qquad
  Sanjay Thasma\textsuperscript{1} \qquad
  Junwei Ma\textsuperscript{1,2} \qquad
  Ali Mostafavi\textsuperscript{1,2,3} \\[4pt]
  \small \textsuperscript{1}Texas A\&M University \quad
  \textsuperscript{2}Resilitix Intelligence LLC \quad
  \textsuperscript{3}Institute for a Disaster Resilient Texas \\[3pt]
  \small \texttt{\{yxiao, yuhsuanho, thasma, jwma, mostafavi\}@tamu.edu} \\[3pt]
  \small \textsuperscript{*}Corresponding author
}

\date{}

\begin{document}

\maketitle

\begin{abstract}
Decision-relevant building damage assessment is critical for prioritizing resources and recovery after a disaster, yet most automated methods either flatten damage into a single severity scale (no damage, minor, major, destroyed) or require paired pre- and post-event imagery that is often unavailable for emerging hazards.  This paper presents Damage-TriageFormer, a single-image, post-event, footprint-conditioned model that produces a damage typology rather than a severity scale.  We contribute: (1) DamageTriage-Bench, a new benchmark built from NOAA Emergency Response Imagery across Hurricane Michael (2018), Hurricane Helene (2024), and the 2025 Los Angeles wildfire complex, with five typology classes that distinguish roof damage from structural damage and, within each, partial from total extent; and (2) Damage-TriageFormer, which extends a DINOv3 ViT-L backbone with a Simple Feature Pyramid for higher-resolution instance pooling, a two-stage gated damage head, and an auxiliary severity-regression objective.  Our model achieves macro F1 of 0.624 on validation and 0.619 on a held-out stratified test set, performing strongest where operational triage needs it most, with per-class F1 of 0.91 and 0.84 on undamaged buildings and total structural collapse, respectively.  While the rare Total Roof Damage class remains difficult due to its limited examples and an inherently ambiguous label boundary, our results show that single-image post-event imagery can support actionable building damage typing, enabling targeted emergency response and resource allocation without a pre-event reference.
\end{abstract}

\section{Introduction}

Rapid post-disaster building assessment is critical for emergency response, recovery planning, and loss estimation.  It is also central to disaster resilience: how quickly and how precisely building-level damage is characterized shapes how affected communities triage unsafe structures, sequence repair and reconstruction, and restore the functionality of their building stock~\cite{bruneau2003seismicResilience,nist2016resilienceGuide}.  Recent advances in remote sensing and deep learning have made large-scale damage mapping increasingly practical~\cite{alshafianIntegratingMachineLearning2024,guAdvancesRapidDamage2024,wangDeepLearningModels2024}, but many benchmark formulations still abstract away distinctions that matter for building-level triage.  In particular, broad severity labels do not explicitly encode whether visible damage is confined to the roof, extends into structural components, or corresponds to total structural collapse; component-level damage assessment remains comparatively rare despite its value for post-hurricane decision making~\cite{roScalableApproachCreate2024}.

Two limitations are especially important for operational use. First, there is a semantic limitation.  Widely used benchmarks, especially xBD~\cite{gupta2019xbd} and its derivatives, typically score buildings with a FEMA-style severity scale: no damage, minor, major, and destroyed.  Severity is useful for aggregate loss estimation, but it collapses the component and mechanism.  A roof that has lost much of its covering and a building whose load-bearing elements have partially failed may both be labeled ``major,'' even though they call for different forms of response.  We therefore study a damage typology that separates roof damage from structural damage and separates partial from total damage within each branch.

Second, there is an operational limitation.  Many damage-assessment methods are bi-temporal: they compare a post-event image with a pre-event image of the same scene.  This is powerful when the paired imagery is available and well registered, but it is brittle in the conditions where rapid assessment matters most.  Matching pre-event imagery may be missing, stale, lower resolution, seasonally different, or geometrically misaligned with emergency-response flyovers; rapid satellite acquisitions may also arrive at off-nadir viewing angles that are underrepresented in standard damage-assessment benchmarks~\cite{diasConditionalExpertsImproved2024}.  A single-image post-event formulation removes this dependency and matches the data that are often available first, including NOAA Emergency Response Imagery collected for post-disaster response~\cite{noaaEmergencyResponseImagery}.

This paper studies footprint-conditioned, post-event-only building damage typology.  Given a single sub-meter post-disaster image tile and building instance masks, the model assigns each building to one of five classes: (0) Undamaged, (1) Partial Roof Damage, (2) Total Roof Damage, (3) Partial Structural Damage, and (4) Total Structural Collapse.  The footprint-conditioned setting is operationally realistic because building-footprint layers are increasingly available from local and state GIS programs as well as curated historical datasets~\cite{lacounty_buildings_dins_2025,ncem_buildingfootprints,baycountygis,uhl2022mtbf33}.  It also isolates the damage-typing problem from building detection, making the headline metric a direct test of whether post-event imagery contains enough visual evidence for building damage typology.

We present Damage-TriageFormer, a model that adapts a DINOv3 ViT-L backbone to this task using a Simple Feature Pyramid for higher-resolution instance pooling, a two-stage damage head that separates any-damage recognition from damaged-class typing, and an auxiliary severity-regression objective.  The model is evaluated on DamageTriage-Bench, an internally curated benchmark built from NOAA Emergency Response Imagery and spanning hurricane and wildfire events.  The central question is deliberately practical: how far can a post-event, footprint-conditioned model go before the absence of a pre-event reference and the ambiguity of fine-grained roof damage become the limiting factors?

Our contributions are as follows:

\textbf{(1) A decision-relevant damage-typology task.}
We formulate a per-building 5-class task that separates roof from structural damage and partial from total extent, replacing the prevailing single-axis FEMA severity scale.  The taxonomy is chosen so that adjacent classes correspond to genuinely different field responses rather than graded versions of the same response.

\textbf{(2) DamageTriage-Bench.}
We introduce a post-event-only, footprint-conditioned benchmark built from NOAA Emergency Response Imagery across three disasters spanning two hazard types (hurricane and wildfire), with stratified building-instance train, validation, and test folds and a typology rubric developed jointly with structural-engineering experts.

\textbf{(3) A reproducible DINOv3 baseline.}
We adapt a DINOv3 ViT-L backbone into Damage-TriageFormer, a baseline recipe that pairs higher-resolution instance pooling and a two-stage gated damage head with long-tailed training; its full configuration (Section~\ref{sec:method}) is reported for reproducibility.

\textbf{(4) A diagnostic error profile.}
Per-class and confusion-level analyses show that the visually salient endpoints (undamaged buildings and total structural collapse) are learnable from a single post-event observation, while rare and visually ambiguous roof-damage categories remain the dominant bottleneck.  This isolates the distinctions that do not require a pre-event reference from those that need additional evidence.

\section{Related Work}

Remote-sensing damage assessment has advanced along three linked design choices: the input modality, the prediction target, and the spatial unit of prediction.  Recent reviews emphasize rapid growth in machine-learning-based damage assessment, while also noting persistent needs around real-time processing, generalization, and operational integration~\cite{alshafianIntegratingMachineLearning2024,guAdvancesRapidDamage2024,wangDeepLearningModels2024}. The present work is positioned at a less common intersection of these choices: post-event-only input, building-instance output, and a component-aware damage typology.  We compare quantitatively against representative methods in Section~\ref{sec:main-results} (Table~\ref{tab:sota_comparison}).

\paragraph{Standard severity-based benchmarks.}
The xBD dataset~\cite{gupta2019xbd,gupta2019creating} established a large-scale benchmark for building damage assessment and helped standardize the bi-temporal setting.  Much of the subsequent literature inherits its 4-class severity target, including two-stream and temporal-fusion models for pre/post-event comparison~\cite{shenBDANetMultiscaleConvolutional2022a,mohammadianSiamixFormerFullytransformerSiamese2023}, hierarchical or high-resolution Transformer architectures~\cite{kaurLargescaleBuildingDamage2023a,chenHRTBDANetworkPostdisaster2024}, state-space change-detection models~\cite{chenChangeMambaRemoteSensing2024}, cross-disaster transfer work~\cite{braik2024automated}, and recent post-only patch classification with frozen DINOv2 and DeiT features~\cite{siva2026patch}.  These studies are central to the field, but the xBD label space is severity-oriented rather than component-aware: it does not explicitly distinguish roof-only damage from structural damage.  Our typology keeps both dimensions: which component appears damaged and how extensive the damage is.

\paragraph{Post-event classifiers.}
Several methods avoid the pre-event dependency and operate on post-event imagery alone.  UHRA-ViT~\cite{singh2023uhra} shows that ultra-high-resolution aerial imagery and semi-supervised pre-training can substantially improve overall accuracy, although the benchmark is hurricane-only and the reported metrics are not directly aligned with a typology task.  EBDC-Net~\cite{hong2022ebdc} similarly studies post-event UAV imagery with a compact CNN, while DamageMap~\cite{galanis2021damagemap} focuses on wildfire damage detection as a binary problem.  Other post-event work targets collapsed-building mapping with limited or no target labels, highlighting the importance of methods that can be adapted rapidly to newly affected regions~\cite{adrianoDevelopingFrameworkRapid2023}. Post-event assessment has also been studied with non-overhead or multimodal evidence, including street-view imagery combined with structured variables such as wind speed, building age, and location context~\cite{xuePosthurricaneBuildingDamage2024a}. These works demonstrate the value of post-event visual evidence, but they generally target binary or severity-style outputs rather than fine-grained roof-versus-structure typing.

\paragraph{Segmentation and detection pipelines.}
Another line of work predicts damage at the pixel level or first detects buildings and then classifies them.  DamageScope~\cite{alshafian2025damagescope} and DeepDamageNet~\cite{alisjahbana2024deepdamagenet} use U-Net-style segmentation architectures, while attention-based high-resolution networks and two-step localization/classification pipelines pursue stronger building localization and damage classification under severe class imbalance~\cite{oludareAttentionbasedTwostreamHighresolution2022,wangBuildingDamageDetection2022}. DamageCAT~\cite{xiao2025damagecat}, our prior work, uses bi-temporal input for typology segmentation on Hurricane Ida.  Detection-plus-classification systems such as EDDA~\cite{hatic2025edda} evaluate a broader end-to-end pipeline that includes localization.  In contrast, our headline evaluation uses instance masks so that the reported macro F1 isolates damage-typing performance rather than conflating it with building detection and instance separation.

\paragraph{Footprint-conditioned evaluation.}
Building footprints are often available before or alongside post-disaster image collection, especially in U.S. settings where county, state, and event-specific GIS programs publish building layers~\cite{lacounty_buildings_dins_2025,ncem_buildingfootprints,baycountygis,uhl2022mtbf33}.  Conditioning on footprints changes the evaluation question: instead of asking a single model to solve localization, instance separation, and damage classification at once, it asks whether post-event imagery contains sufficient evidence to type each known building.  This distinction matters because xBD-style damage models can over-rely on surrounding damage context rather than the target building itself~\cite{melamedUncoveringBiasBuilding2024}. This makes the setting particularly useful for studying fine-grained typology, where confusion among roof damage, structural damage, and collapse can be obscured by detection errors in a fully end-to-end pipeline.

\paragraph{Pretrained visual features.}
Recent damage-assessment work has begun to test whether large pretrained vision backbones transfer to post-disaster imagery. Remote-sensing foundation-model work, such as SatMAE, shows that pretraining can exploit large volumes of unlabeled satellite imagery and transfer to downstream geospatial tasks~\cite{congSatMAEPretrainingTransformers2023}. For damage assessment specifically, self-supervised xBD studies and vision-foundation-model change-detection frameworks suggest that large pretrained representations can improve robustness when target labels are scarce or domains shift~\cite{xiaSelfSupervisedLearningBuilding2022,ahnGeneralizableDisasterDamage2025}. Patch-based studies using frozen DINOv2 and DeiT features suggest that foundation-style representations are useful even without pre-event imagery~\cite{siva2026patch}.  We build on this direction with DINOv3 ViT-L (\texttt{facebook/dinov3-vitl16-pretrain-lvd1689m}), but adapt it to building damage typology by adding higher-resolution instance pooling and a two-stage damage head rather than treating each building crop as an independent image patch.

\section{Dataset: DamageTriage-Bench}
\label{sec:dataset}

\paragraph{Source.}
DamageTriage-Bench is internally curated from NOAA Emergency Response Imagery (ERI), a high-resolution remotely sensed imagery product collected by NOAA's National Geodetic Survey for post-disaster response~\cite{noaaEmergencyResponseImagery}.  The dataset covers three disasters across two hazard types: Hurricane Michael (2018), Hurricane Helene (2024), and the 2025 Los Angeles Palisades and Eaton wildfire complex.  The ERI catalog reports 0.3--0.5 m imagery. Candidate building footprints were assembled from event-appropriate public GIS sources: Los Angeles County building polygons with CAL FIRE Damage Inspection (DINS) attributes for the Palisades and Eaton fires, North Carolina Emergency Management/NC OneMap footprints for Hurricane Helene, and Bay County GIS layers for Hurricane Michael~\cite{lacounty_buildings_dins_2025,ncem_buildingfootprints,baycountygis}. Damage labels were then manually annotated at the instance level by trained annotators following a damage-typology rubric jointly developed with structural-engineering domain experts; each polygon is assigned exactly one of the five typology classes.  All scenes are tiled to non-overlapping $1024 \times 1024$ patches, and tiles containing no building footprint are discarded.

\paragraph{Damage-typology rubric.}
For damaged buildings, labels are assigned using a two-step rule (Table~\ref{tab:rubric}).  Annotators first determine whether visible damage extends below the roof surface into the structural components of the building.  Damage confined to the roof surface is labeled as roof damage; damage extending below the roof is labeled as structural damage.  Within each branch, a 50\% affected-area threshold separates partial from total damage, with area estimated relative to the visible roof or building footprint.

\begin{table}[!htbp]\centering
\caption{Damage-typology class definitions used for annotation.  Each building instance receives exactly one class by a two-step rule: component first (roof vs.\ structural), then a 50\% affected-area threshold for partial vs.\ total.}
\label{tab:rubric}
\small
\setlength{\tabcolsep}{6pt}
\renewcommand{\arraystretch}{1.2}
\begin{tabularx}{\linewidth}{@{}>{\rr}p{0.26\linewidth} >{\rr}X@{}}
\toprule
\textbf{Class} & \textbf{Operational definition} \\
\midrule
\textbf{0. Undamaged} &
No visible roof or structural damage. \\
\addlinespace
\textbf{1. Partial Roof Damage} &
Visible damage is limited to the roof surface and affects less than 50\% of the roof area; the building is treated as habitable under the annotation rubric. \\
\addlinespace
\textbf{2. Total Roof Damage} &
Visible damage is limited to the roof surface but affects more than 50\% of the roof area. \\
\addlinespace
\textbf{3. Partial Structural Damage} &
Visible damage involves structural components below the roof, the building is treated as not habitable under the annotation rubric, and the affected area is less than 50\%. \\
\addlinespace
\textbf{4. Total Structural Collapse} &
Visible structural damage affects more than 50\% of the building footprint. \\
\bottomrule
\end{tabularx}
\end{table}

\paragraph{Typology-to-decision mapping.}
Table~\ref{tab:typology_decision} makes the decision-relevant framing concrete: each class is paired with an operational interpretation, an example downstream decision, and the dominant risk if the prediction is wrong.  This mapping is the operational justification for separating roof from structural damage and partial from total extent rather than reporting a single severity score.

\begin{table}[!htbp]\centering
\caption{Typology-to-decision mapping.  Each class is connected to an operational interpretation, a representative decision use, and the dominant risk under misclassification.}
\label{tab:typology_decision}
\small
\setlength{\tabcolsep}{6pt}
\renewcommand{\arraystretch}{1.2}
\begin{tabularx}{\linewidth}{@{}>{\rr}p{0.17\linewidth} Y{0.8} Y{1.4} Y{0.8}@{}}
\toprule
\textbf{Class} & \textbf{Operational meaning} & \textbf{Potential decision use} & \textbf{Decision risk} \\
\midrule
\textbf{Undamaged} & No visible roof or structural damage. & Lower-priority field inspection; area-level situational awareness; sampling-based validation. & False negative risk if subtle damage is missed. \\
\midrule
\textbf{Partial Roof Damage} & Localized roof-surface damage. & Roof-repair triage; insurance screening; temporary weatherproofing prioritization. & May be under-triaged as undamaged or over-triaged as structural damage. \\
\midrule
\textbf{Total Roof Damage} & Roof-surface damage affecting most of the roof area. & Urgent roof repair/claim triage; habitability screening; manual review under low confidence. & Rare and ambiguous; false negatives delay needed repair support. \\
\midrule
\textbf{Partial Structural Damage} & Damage extends below the roof into structural components, partial extent. & Structural engineering inspection; habitability review; safety-focused field verification. & Under-triage creates safety risk; over-triage consumes scarce inspection capacity. \\
\midrule
\textbf{Total Structural Collapse} & Visible structural damage across most of the footprint. & Highest-priority safety response; exclusion/red-tag screening; emergency-management situational awareness. & False negatives carry the highest operational consequence. \\
\bottomrule
\end{tabularx}
\end{table}

\paragraph{Splits.}
We use a stratified per-event split: within each of the twelve acquisition sub-events spanning the three disasters, tiles are partitioned into 70\% train, 15\% validation, and 15\% test splits, so that every split has the same per-sub-event proportions as the full dataset. Tiles are the split unit; events are too few to support a per-event hold-out without losing data diversity.  After tile filtering this yields 5{,}229 train, 1{,}120 validation, and 1{,}123 test tiles.  This protocol measures within-event generalization; event-held-out and region-held-out variants are flagged in Section~\ref{sec:discussion} as the next benchmark steps.

\paragraph{Annotation quality.}
Borderline roof/structural cases were flagged and reviewed against the two-step rule (component first, then extent), and the dataset deliberately retains common boundary cases (for example, extensive roof loss exposing rafters that can be argued for either Total Roof Damage or Partial Structural Damage) rather than discarding ambiguity.

\paragraph{Instance class supports.}
Table~\ref{tab:supports} reports per-class instance counts in the stratified split; these counts are the denominators for all reported recall figures in Section~\ref{sec:results}.

\begin{table}[!htbp]\centering
\caption{Per-class building-instance counts in the validation and test folds of the stratified split, with each fold's percentage composition.  These supports are the denominators for the per-class recall figures in Section~\ref{sec:results}.}
\label{tab:supports}
\small
\begin{tabular}{lrrrr}
\toprule
\textbf{Class} & \textbf{Val.} & \textbf{Val.\ \%} & \textbf{Test} & \textbf{Test \%} \\
\midrule
0. Undamaged                  & 9{,}288 & 73.1 & 9{,}346 & 74.6 \\
1. Partial Roof Damage        & 1{,}657 & 13.0 & 1{,}561 & 12.5 \\
2. Total Roof Damage          &    165  &  1.3 &    145  &  1.2 \\
3. Partial Structural Damage  &    543  &  4.3 &    471  &  3.8 \\
4. Total Structural Collapse  & 1{,}062 &  8.3 &    999  &  8.0 \\
\midrule
\textbf{Total}                & 12{,}715 & 100.0 & 12{,}522 & 100.0 \\
\bottomrule
\end{tabular}
\end{table}

\noindent The training split follows the same long-tail distribution with the remaining roughly 60{,}000 instances.  This imbalance motivates the macro-F1 reporting and per-class analysis in Section~\ref{sec:results}.

\section{Method}
\label{sec:method}

\subsection{Architecture overview}

\begin{figure}[!htbp]
  \centering
  \includegraphics[width=\linewidth]{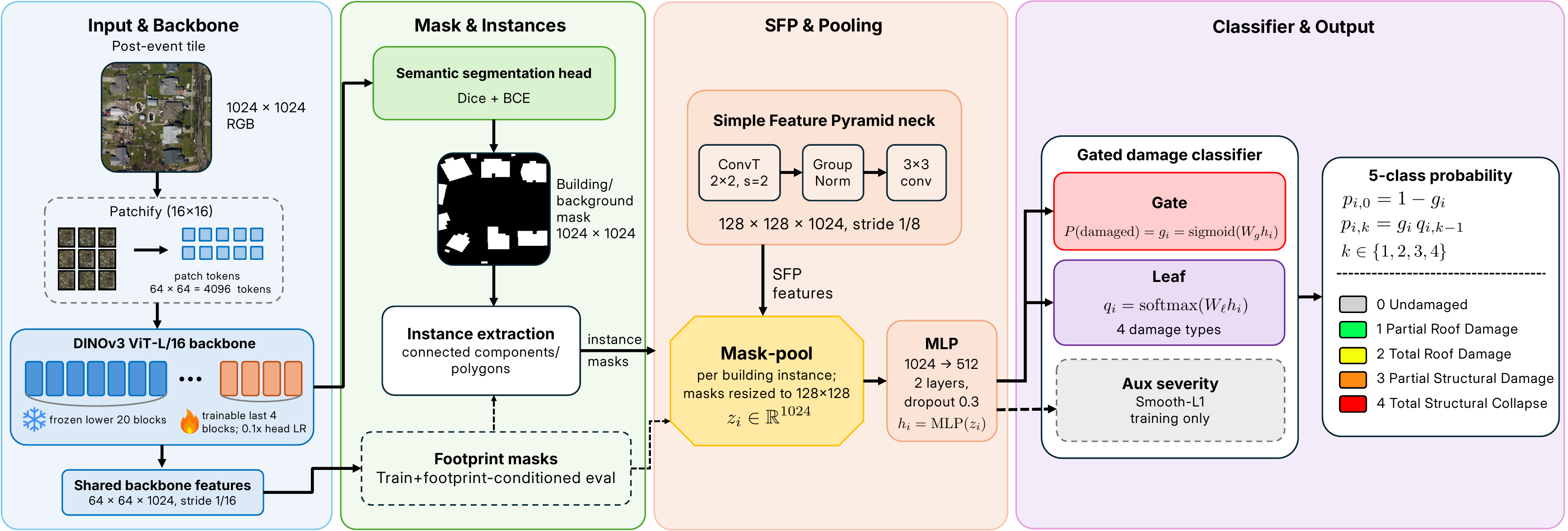}
  \caption{Overview of Damage-TriageFormer.  A post-event image tile is encoded with a partially fine-tuned DINOv3 ViT-L/16 backbone. Footprint instance masks define the buildings used for instance mask pooling over the SFP feature map.  The pooled feature $z_i$ is projected to an instance embedding $h_i$, from which a gate head predicts the probability of any damage and a leaf head predicts the conditional 4-way damage type.  The final 5-class distribution is formed as $\hat{p}_{i,0}=1-g_i$ and $\hat{p}_{i,k}=g_i q_{i,k-1}$ for $k\in\{1,2,3,4\}$.}
  \label{fig:overview}
\end{figure}

Given a $1024\times 1024$ post-event RGB tile $\mathbf{x}$ and a set of building instance masks, the model produces for each building instance a 5-way damage-class probability vector $\hat{\mathbf{p}}\in\Delta^4$. Figure~\ref{fig:overview} summarizes the full pipeline.

The architecture comprises a DINOv3 backbone, an instance feature-pooling stage, and a two-stage damage head, with an auxiliary segmentation branch for building localization.  We describe each below.

\paragraph{Backbone.}
A DINOv3 ViT-L/16 encoder \cite{simeoniDINOv32025} pre-trained on the LVD-1689M dataset.  The first 20 transformer blocks are frozen except for LayerNorm parameters; the last 4 are fine-tuned with a learning-rate multiplier of $0.1\times$ relative to the heads.

\paragraph{Segmentation head.}
A lightweight up-sampling decoder maps the $64 \times 64$ patch embeddings back to a $1024 \times 1024$ binary building-vs-background segmentation map, trained with a Dice + binary cross-entropy combination as an auxiliary localization task.

\paragraph{From segmentation to instance masks.}
Ground-truth instance polygons are rasterized to binary masks $\{m_i\}$ and used directly to pool features for each building instance for the headline damage metrics; the segmentation head is supervised with the union of these masks.  When external footprints are unavailable, individual instance masks can instead be recovered by thresholding the predicted sigmoid map at $0.5$ and running 8-connectivity connected-component labeling, dropping components whose pixel area falls below a small minimum area threshold of 30 pixels.

\paragraph{Feature pooling.}
The codebase exposes three optional feature-enrichment modules before pooling; we describe all three for completeness, but the final recipe (Section~\ref{sec:final-config}) uses only the third (SFP) and disables the other two. Atrous Spatial Pyramid Pooling (ASPP) runs four parallel convolutions with dilation rates $\{1, 1, 2, 4\}$ and fuses the result, capturing different receptive-field sizes at the same spatial resolution.  A learned spatial attention module produces a sigmoid attention map and replaces the plain mask average.  Both modules operate at the native ViT stride $1{:}16$.  They aggregate context while leaving the instance feature map at the same spatial resolution.  In pilot experiments at our scale, neither module improved macro-F1, and both are disabled in the reported configuration.

\paragraph{Simple Feature Pyramid (SFP).}
For higher spatial resolution, we optionally apply a Simple Feature Pyramid head \cite{liExploringPlainVision2022} between the backbone and the instance pool.  Concretely, the $64 \times 64$ feature map is passed through $\log_2(s_\text{src}/s_\text{tgt})$ transposed $2 \times 2$ convolutions interleaved with GroupNorm and GELU, plus a $3 \times 3$ refinement conv at the target resolution.  Setting target stride $s_\text{tgt}=8$ doubles the spatial resolution to $128 \times 128$; $s_\text{tgt}=4$ quadruples it to $256 \times 256$.  In the reported recipe, instance pooling is a plain mask-weighted average over the SFP-upsampled feature map (ASPP and attention pooling disabled, as above).  The segmentation head still consumes the original $1{:}16$ features (it has its own upsampling decoder); only the damage head benefits from SFP.

\paragraph{Two-stage damage head.}
For each building instance mask $m_i$, we then mask-pool the SFP-upsampled backbone embedding to a single $D$-dim vector $z_i$.  A shared MLP projects this pooled feature to an instance embedding $h_i$.  The head first predicts whether the building is damaged, then predicts the damage type conditional on damage:

\begin{align}
a_i        &= f_g(h_i), \qquad
\boldsymbol{b}_i = f_\ell(h_i) \\
g_i        &= \sigma(a_i) \in [0,1]
                       && \text{(probability of any damage)} \\
\boldsymbol{q}_i &= \mathrm{softmax}(\boldsymbol{b}_i) \in \Delta^3
                       && \text{(4-way damage type)}
\end{align}

Here, $f_g$ and $f_\ell$ are the small MLP gate and leaf heads implemented in the model code.

The 5-class probability is reconstructed at inference as
\begin{equation}
\hat{p}_{i,k} =
\begin{cases}
1 - g_i, & k = 0\ \text{(undamaged)}\\
g_i \cdot \boldsymbol{q}_{i,k-1}, & k \in \{1,2,3,4\}.
\end{cases}
\end{equation}

\subsection{Training loss and class imbalance}

The reported objective combines the following components:

\begin{enumerate}
  \item A gated damage loss
        $\mathcal{L}=\mathcal{L}_\text{seg}
        +\lambda_g\mathcal{L}_\text{gate}
        +\lambda_\ell\mathcal{L}_\text{leaf}
        +\lambda_s\mathcal{L}_\text{sev}$, with
        $(\lambda_g,\lambda_\ell,\lambda_s)=(1.0,2.0,0.5)$ in the
        reported run.  $\mathcal{L}_\text{seg}$ is Dice + BCE,
        $\mathcal{L}_\text{gate}$ is binary BCE over damaged versus
        undamaged buildings, and $\mathcal{L}_\text{leaf}$ is
        cross-entropy over the four damaged classes, computed only
        for damaged instances.
  \item Class weighting on the leaf classifier: the 4-way damaged
        loss uses inverse-square-root 5-class weights
        $w_c \propto 1/\sqrt{f_c}$, indexed by the original class
        label.  The reported configuration leaves the binary gate
        BCE unweighted because no separate gate-class weights are
        enabled.
  \item Logit adjustment~\cite{menonLongtailLearningLogit2021} during
        training only ($\tau=1.0$).  The gate logit is shifted by
        $\tau\log(\pi_d/(1-\pi_d))$, where $\pi_d$ is the damaged
        prior, and the damaged-class leaf logits are shifted by
        $\tau\log\pi_{c|d}$ using priors conditional on the building
        being damaged.  No logit adjustment is applied at inference.
  \item Auxiliary severity-regression head: a small MLP
        head that predicts a continuous severity scalar
        $s_i \in [0, 1]$ for each instance via a smooth-L1 loss
        against per-class severity targets
        $\{0.0, 0.3, 0.7, 0.5, 1.0\}$ for the 5 classes.  The aux
        head shares the pooled instance feature with the
        binary and damage-type heads and is used only during
        training.  Its role is to encourage the shared embedding
        to encode an interpretable severity axis along which the
        four damage classes are ordered.
\end{enumerate}

\subsection{Final configuration}
\label{sec:final-config}

For reproducibility, the configuration whose results are reported in Section~\ref{sec:main-results} fixes the modules above as follows: (i) DINOv3 ViT-L/16 backbone with the last 4 transformer blocks and LayerNorm parameters fine-tuned at a $0.1\times$ LR multiplier; (ii) a Simple Feature Pyramid head at target stride 8 (1/8 feature map, $128\times128$ at a $1024$-pixel input); (iii) instance pooling as a plain mask-weighted average over the SFP-upsampled features (with ASPP and learned spatial attention pooling disabled); (iv) the two-stage damage head paired with the auxiliary severity-regression head described above; (v) logit adjustment at $\tau{=}1.0$ applied to the gate and leaf logits during training only; (vi) inverse-square-root class weighting on the leaf cross-entropy, no focal loss, label smoothing $0.1$, and EMA weight averaging with decay $0.9995$. Optimization hyperparameters (LR, batch size, epochs, schedule) are reported in Section~\ref{sec:results}.

\section{Experiments}
\label{sec:results}

\subsection{Setup}

\paragraph{Training.}
We train for 30 epochs with AdamW (LR $5\times10^{-5}$ on heads, $5\times10^{-6}$ on the unfrozen backbone blocks), per-GPU batch size 2 with gradient accumulation $\times 8$ (effective batch 32 across 2 GPUs), label smoothing $0.1$, EMA model averaging (decay 0.9995), and a 2-epoch linear warm-up followed by cosine decay. Hardware: 2$\times$A100 40GB GPUs with PyTorch DistributedDataParallel on the TAMU FASTER cluster.  Validation runs every epoch; the checkpoint is selected by validation macro F1 in the footprint-conditioned (mask-conditioned) setting, which we use throughout the pipeline for both training-time validation and headline reporting unless explicitly stated otherwise.

\paragraph{Splits and roles.}
We use the stratified per-event split (5{,}229 train, 1{,}120 validation, and 1{,}123 test tiles) described in Section~\ref{sec:dataset}.  Model selection (best epoch by footprint-conditioned macro F1) is done on the validation fold.  The test fold is never read during training, model selection, or any hyperparameter tuning, and is consulted only once to produce the final numbers reported in Section~\ref{sec:main-results}.

\subsection{Main results}
\label{sec:main-results}

Our model achieves a macro F1 of 0.624 on the validation fold and 0.619 on the held-out test fold of the stratified split.  Performance is strongly class-dependent (Table~\ref{tab:perclass}): the two visually distinct endpoints are the easiest, with Undamaged buildings reaching F1 around 0.91 and Total Structural Collapse around 0.84 on both folds, while the two partial classes sit in the middle (Partial Roof Damage 0.47--0.52, Partial Structural Damage 0.53--0.55) and Total Roof Damage is the weakest by a wide margin at F1 around 0.33.  Validation and test F1 agree to within a few points for every class, the largest gap being Partial Roof Damage (0.515 versus 0.472), which indicates stable generalization across the stratified folds rather than overfitting to the validation set.  Per-class supports are listed in Table~\ref{tab:supports}.

\begin{table}[!htbp]\centering
\caption{Per-class and macro F1 on the validation and test folds.  Per-class supports are reported in Table~\ref{tab:supports}.}
\label{tab:perclass}
\small
\begin{tabular}{lrr}
\toprule
\textbf{Class} & \textbf{F1\textsubscript{val}} & \textbf{F1\textsubscript{test}} \\
\midrule
Undamaged                   & 0.907 & 0.906 \\
Partial Roof Damage         & 0.515 & 0.472 \\
Total Roof Damage           & 0.336 & 0.326 \\
Partial Structural Damage   & 0.526 & 0.554 \\
Total Structural Collapse   & 0.835 & 0.836 \\
\midrule
\textbf{Macro}              & 0.624 & 0.619 \\
\bottomrule
\end{tabular}
\end{table}

\paragraph{Comparison with prior work.}
Table~\ref{tab:sota_comparison} places these numbers in context, though the rows should be read as a landscape rather than a strict leaderboard: they differ in hazard scope, sensor platform, spatial output unit, label taxonomy, and whether the model sees pre-event imagery.  Several methods report higher numbers under different task definitions, including binary damage, ultra-high-resolution aerial imagery, or different class taxonomies; Sat-ViT-100~\cite{singh2023uhra}, for example, reaches a macro F1 of 0.72 on hurricane-only xBD crops with a 5-class FEMA scale.  Our row reports footprint-conditioned building-level metrics under a narrower combination of design choices: a fine-grained 5-class roof and structural typology, multi-disaster wildfire and hurricane scope, single-image post-event input without a pre-event reference or change-detection branch, and a reported macro F1.  Overall accuracy is included for continuity with prior work, but it is heavily shaped by the roughly 74\% Undamaged majority; macro F1 and per-class F1 are the more informative metrics for this long-tailed typology setting.

\begin{table*}[!ht]
\centering
\caption{Contextual comparison of post-event-only building damage assessment methods, grouped by task: per-building classification (top), per-pixel segmentation (middle), and detection with classification (bottom).  Task definitions, sensors, output units, and class taxonomies differ across rows, so the table is intended as context rather than a direct leaderboard.  ``Disaster Scope'' is the number of unique events with distinct hazard types in parentheses; ``---'' marks unreported metrics.}
\label{tab:sota_comparison}
\resizebox{\textwidth}{!}{%
\begin{tabular}{lcllclccc}
\toprule
\textbf{Method} & \textbf{Year} & \textbf{Architecture} & \textbf{Dataset} & \textbf{Classes} & \textbf{Disaster Scope} & \textbf{OA (\%)} & \textbf{Macro F1} & \textbf{Post-Only} \\
\midrule
\multicolumn{9}{l}{\textbf{Per-building classification}} \\
\midrule
\shortstack[l]{\textbf{Damage-TriageFormer}\\\textbf{(Ours)}} & \textbf{2026} & \shortstack[l]{\textbf{DINOv3 ViT-L + SFP}\\\textbf{+ two-stage head}}        & \textbf{DamageTriage-Bench}             & \textbf{5 (typology)} & \textbf{3 (2)}                          & \textbf{82.3} & \textbf{0.619} & \textbf{Yes} \\
xBD Baseline\textsuperscript{a}  & 2019 & ResNet50                      & xBD                         & 4 (severity)          & 19 (6)                                  & $\sim$75      & $\sim$0.27      & Yes \\
Sat-CNN-100 \cite{singh2023uhra}   & 2023 & CNN                           & xBD satellite crops         & 5 (FEMA)              & $\leq$6 (1)                             & 55            & 0.54            & Yes \\
Sat-ViT-100 \cite{singh2023uhra}   & 2023 & ViT                           & xBD satellite crops         & 5 (FEMA)              & $\leq$6 (1)                             & 73            & 0.72            & Yes \\
UHRA-ViT-100 \cite{singh2023uhra} & 2023 & ViT + semi-supervised & UHRA aerial crops   & 5 (FEMA)              & 1 (1)                                   & 88            & ---             & Yes \\
EBDC-Net\textsuperscript{b} \cite{hong2022ebdc}       & 2022 & Custom CNN            & Yangbi/Ludian/Yushu UAV     & 4 (severity)          & 3 (1)                                   & 77.5          & ---             & Yes \\
Braik \& Koliou\textsuperscript{c} \cite{braik2024automated} & 2024 & CNN (xBD-pretrained) & xBD to Galveston & 4 (severity) & 19 to 1 (1)                    & 93            & ---             & Yes \\
Patch-ViT\textsuperscript{d} \cite{siva2026patch}     & 2026 & DINOv2-S (frozen)     & xBD post-only crops         & 4 (severity)          & $\leq$19 ($\leq$6)                      & ---           & ---             & Yes \\
Patch-ViT\textsuperscript{d} \cite{siva2026patch}     & 2026 & DeiT (frozen)         & xBD post-only crops         & 4 (severity)          & $\leq$19 ($\leq$6)                      & ---           & ---             & Yes \\
DamageMap \cite{galanis2021damagemap}                  & 2021 & CNN                   & Camp/Carr Fire              & 2 (binary)            & 7 (1)                                   & 92--98        & ---             & Yes \\
\midrule
\multicolumn{9}{l}{\textbf{Per-pixel segmentation}} \\
\midrule
DamageScope\textsuperscript{e} \cite{alshafian2025damagescope}  & 2025 & U-Net + FastViT-MA36  & xBD (post-only)     & 4 (severity)          & 19 (6)                                  & ---           & 0.598           & Yes \\
DeepDamageNet\textsuperscript{f} \cite{alisjahbana2024deepdamagenet} & 2024 & U-Net + ResNet50 & xBD                & 4 (severity)          & 19 (6)                                  & ---           & ---             & Partial \\
DamageCAT\textsuperscript{g} \cite{xiao2025damagecat}            & 2025 & Hier.\ U-Net Trans.  & BD-TypoSAT (Ida)    & 4 (typology)          & 1 (1)                                   & ---           & 0.8835          & No \\
\midrule
\multicolumn{9}{l}{\textbf{Detection with classification}} \\
\midrule
EDDA two-stage\textsuperscript{h} \cite{hatic2025edda}         & 2025 & RTMDet + CCT          & EDDA (Mozambique)   & 3 (severity)          & 1 (1)                                   & ---           & 0.863           & Yes \\
\bottomrule
\end{tabular}%
}

\vspace{4pt}
\raggedright
\footnotesize

\textsuperscript{a} Roughly 75\% OA is a trivial no-damage baseline under xBD imbalance (not formally reported); weighted F1 0.2654~\cite{gupta2019xbd} used as a macro-F1 proxy. \\
\textsuperscript{b} Only OA reported (77.5\% for the 4-class grouping); no macro F1. \\
\textsuperscript{c} 93\% OA; weighted F1 0.88 on xBD, 0.86 after Galveston fine-tuning; macro F1 not reported. \\
\textsuperscript{d} Per-class metrics only; no headline macro F1. \\
\textsuperscript{e} Macro of reported per-class F1 is 0.598 (per-pixel, not per-building). \\
\textsuperscript{f} xView2 combined metric (0.66), not standard macro F1; segmentation uses pre-event imagery, classification post-event. \\
\textsuperscript{g} Bi-temporal per-pixel segmentation; 4 typology classes, no undamaged class. \\
\textsuperscript{h} Overall F1 over 3 UAV classes (destroyed 0.49); platform and taxonomy differ from ours.

\end{table*}

\begin{figure}[!htbp]
  \centering
  \includegraphics[width=0.6\linewidth]{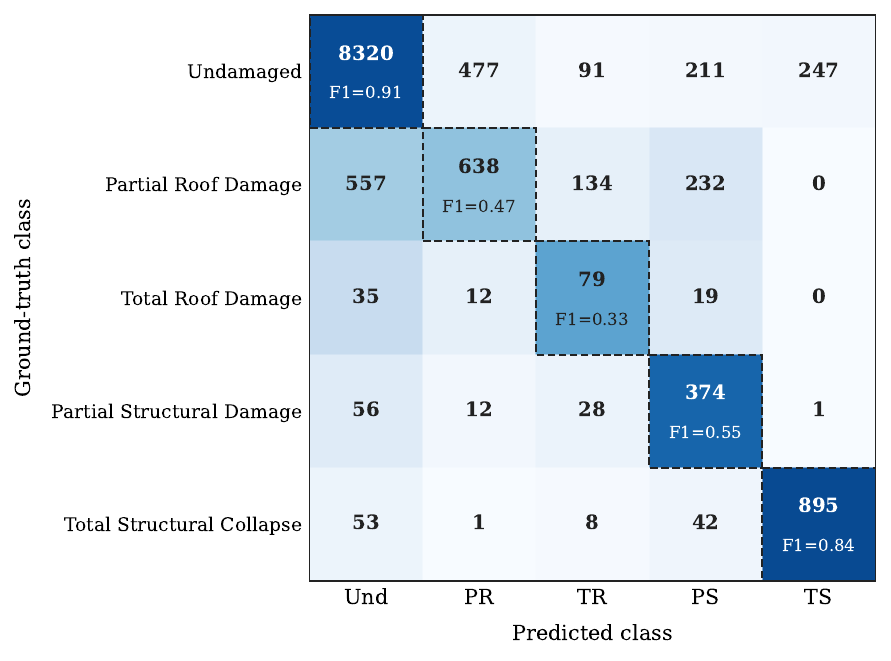}
  \caption{Confusion matrix on the test fold (1{,}123 tiles, footprint-conditioned setting).  Cells show raw instance counts; diagonal cells additionally show per-class F1.}
  \label{fig:confusion}
\end{figure}
\paragraph{Confusion matrix.}
Figure~\ref{fig:confusion} shows the full 5$\times$5 confusion matrix on the test fold in the footprint-conditioned setting, so row totals equal the per-class GT supports in Table~\ref{tab:supports}.  Total Roof Damage instances are confused with Undamaged (24\%), Partial Structural Damage (13\%), and Partial Roof Damage (8\%); 55\% are correctly classified.  Total Structural Collapse, in contrast, reaches 90\% recall.  Undamaged (89\%) and Partial Structural Damage (79\%) also dominate the diagonal; the persistent off-diagonal mass is between the two roof classes and in Partial Roof Damage instances predicted as Undamaged.  The model identifies a much larger share of Total Roof Damage instances than a naive classifier on this imbalanced 5-way problem would suggest, although rare-class precision remains lower than for the head classes.

\begin{figure}[!htbp]
  \centering
  \includegraphics[width=0.96\linewidth]{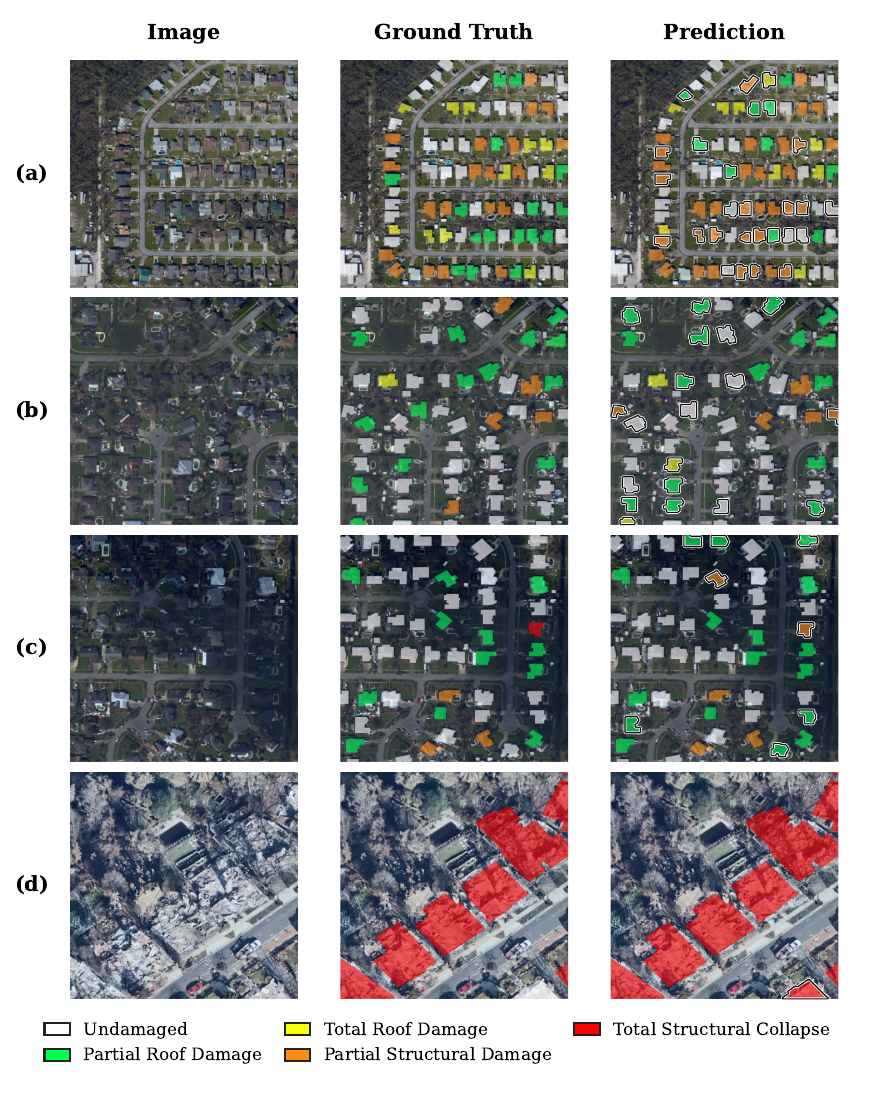}
  \caption{Qualitative predictions on DamageTriage-Bench test tiles.  Each row shows the original tile (left), the ground-truth damage typology overlay (middle), and the model's prediction overlay (right).  Color key: white = Undamaged, green = Partial Roof Damage, yellow = Total Roof Damage, orange = Partial Structural Damage, and red = Total Structural Collapse.  Black-and-white footprint outlines in the prediction column mark ground-truth/prediction disagreements.}
  \label{fig:qualitative}
\end{figure}

\paragraph{Qualitative examples.}
Figure~\ref{fig:qualitative} shows four densely populated DamageTriage-Bench test tiles on which the model produces accurate predictions, picked from a pool of 400 inference triptychs to span hazard type and damage typology: (a)--(c) three Hurricane Michael suburban tiles showing mixed Partial Roof Damage, Partial Structural Damage, and isolated Total Roof Damage and Total Structural Collapse instances against a dominant Undamaged background, and (d) a wildfire neighborhood in which the great majority of buildings are totally collapsed.  Each tile contains tens to hundreds of building instances.  In every row, the prediction overlay closely tracks the ground-truth overlay across both spatial distribution and damage typology; the small footprint outlines in the prediction column mark the residual disagreements.

\subsection{Training dynamics}
\label{sec:training-dynamics}

\begin{figure}[!htbp]
  \centering
  \includegraphics[width=\linewidth]{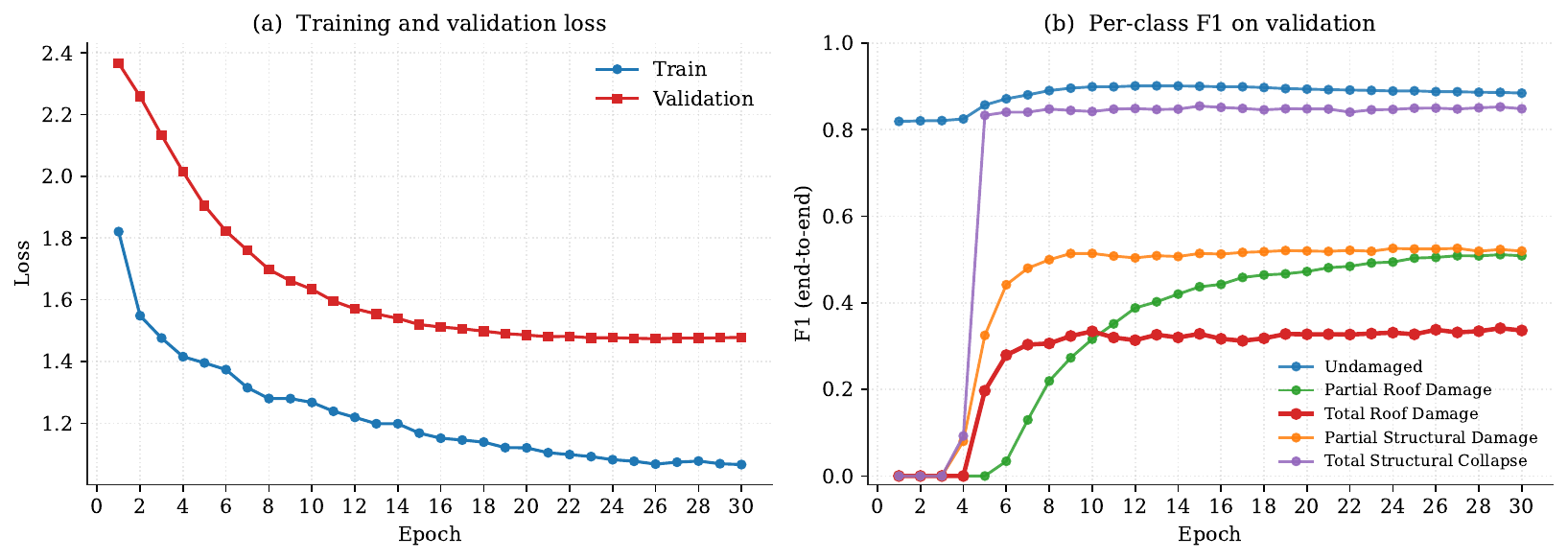}
  \caption{Training dynamics over 30 epochs.  (a)~Train (blue) and validation (red) loss both decrease monotonically with no divergence between curves.  (b)~Per-class F1 on validation: Total Roof Damage (red) reaches its plateau at F1 around 0.33 by epoch~7 and shows no further gain over the remaining roughly 23 epochs, while every other class continues to improve.}
  \label{fig:training_curves}
\end{figure}

Figure~\ref{fig:training_curves} shows the training trajectory. Train and validation loss decrease monotonically over all 30 epochs (panel~a), with the validation curve tracking training without divergence.  The per-class F1 curves (panel~b) make the rare-class behavior visible: Total Roof Damage reaches F1 around 0.33 by epoch~7 and then plateaus, while every other class continues to climb.  This per-class plateau, which persists despite a steadily shrinking loss, suggests that training longer alone is unlikely to remove the rare-class ceiling.

\section{Discussion}
\label{sec:discussion}

\paragraph{What post-event imagery alone can resolve.}
Footprint-conditioned post-event imagery carries more decision-relevant signal than a binary damaged/undamaged framing would suggest.  On the test fold, the model reaches high F1 on both undamaged buildings and Total Structural Collapse, and Partial Structural Damage sits well above the rare Total Roof Damage class.  In practice, the model's strength is not fine-grained roof extent estimation but the separation of buildings that can likely be deprioritized from those that need urgent structural attention.  This separation is already a useful operating regime for emergency screening when pre-event imagery is unavailable, misregistered, or collected under very different viewing conditions.

\paragraph{Implications for resilience and recovery.}
For civil engineering and community resilience, the value of this output is that it is keyed to the building decisions that drive recovery rather than to an abstract severity score.  Separating roof damage from structural damage maps onto the distinction between weatherproofing or repair needs and the life-safety, load-bearing concerns that govern structural inspection, red-tagging, and habitability judgments~\cite{atc2005windstormFloodEval}.  Because predictions are attached to individual building footprints, they can be aggregated into the building-stock functionality and restoration-time estimates that resilience and loss-estimation frameworks depend on~\cite{cimellaro2010resilienceQuantification,femaHazusHurricane}, giving responders an early, spatially explicit picture of where engineering capacity and recovery effort should be concentrated.

\paragraph{Why footprint conditioning matters.}
Conditioning on footprints separates two problems that a single end-to-end score would conflate: finding buildings and assigning damage typology.  Building-footprint layers are widely maintained by local and state GIS programs, so this is not an artificial decoupling; it asks how far post-event visual evidence can go once the building support is known. The auxiliary segmentation head indicates what remains before a fully footprint-free deployment is possible.

\paragraph{The bottleneck.}
Total Roof Damage remains the hardest class (F1 around 0.33 on both folds).  As the confusion matrix shows (Figure~\ref{fig:confusion}), its errors split into two distinct failure modes: a detection failure, where the building is read as Undamaged, and a fine-grained typing failure, where it is confused with the adjacent partial roof and structural classes.  Because SFP already gives the damage head $4\times$ more feature cells (2$\times$ per side), feature resolution alone is unlikely to explain the ceiling.  Three structural factors compound: small support (only $n{=}145$ test instances), an inherently difficult partial/total roof boundary at sub-meter resolution, and a taxonomy whose adjacent classes form a continuum more than isolated visual concepts.

\paragraph{Limitations.}
The test fold is a 15\% stratified sample within each event, which tests tile-level generalization under a realistic data mix but not event-held-out or region-held-out generalization.  Rare-class supports are small, so tail F1 is sensitive to a handful of instance-level errors; combined with the single-seed training run, small numerical differences should be interpreted cautiously.  Finally, the headline metric assumes available building instance masks; the auxiliary segmentation head shows a path toward footprint-free deployment, but is not part of the headline score, and a fully deployable system must jointly evaluate detection, instance separation, and damage typing.  Beyond macro F1, future evaluation should also report decision-sensitive error costs, calibration (reliability diagrams, expected calibration error), and abstention behavior on ambiguous roof-damage cases.

\section{Conclusion}

We presented Damage-TriageFormer, a footprint-conditioned model for post-event damage typology.  The work combines a component-aware roof/structural taxonomy, single-image post-event input, and a DINOv3-based model for settings where building footprints are available.  On DamageTriage-Bench, the model reaches macro F1 of 0.624 on validation and 0.619 on a stratified test fold.

The central conclusion is that post-event-only damage typology is not merely a weakened substitute for change detection.  When building support is known, post-event aerial imagery contains enough evidence to resolve several distinctions that matter for triage, especially the separation of undamaged buildings from total structural collapse.  This makes the footprint-conditioned setting a useful operational target and a useful scientific diagnostic: it tests semantic damage recognition without conflating it with building localization.  For civil and resilience applications, the practical implication is that a single post-event flyover can already yield a building-level, component-aware damage layer suited to structural triage and recovery prioritization, rather than only an aggregate severity map.

At the same time, the results define a clear ceiling for the current system.  Total Roof Damage remains the limiting class, and its errors cannot be attributed to one module alone.  They arise from the interaction of class imbalance, subtle visual evidence, and a taxonomy whose roof extent boundary is inherently difficult at sub-meter resolution.  This is where additional data, adjudicated labels, and uncertainty-aware evaluation are likely to matter as much as architecture.

The next step is therefore broader than another architecture swap.  A deployable post-event typology system will need stronger event-held-out testing, explicit treatment of ambiguous roof-damage labels, better rare class coverage, and reliable integration with footprint-free detection when building masks are not available.  DamageTriage-Bench and Damage-TriageFormer provide a starting point for that agenda: a concrete task, a reproducible footprint-conditioned baseline, and an error profile that makes the next bottlenecks visible.

\section*{Data and Code Availability}
The training and inference code for Damage-TriageFormer is available on \href{https://github.com/YimingXiao98/Damage-TriageFormer}{GitHub} under the MIT License. The DamageTriage-Bench benchmark, comprising the post-event image tiles, building-instance damage-typology masks, and stratified train/validation/test splits, is available on \href{https://huggingface.co/datasets/Ymx1025/DamageTriage-Bench}{Hugging Face} under the CC-BY-NC-4.0 License, subject to the redistribution terms of the underlying NOAA Emergency Response Imagery and the source building-footprint layers.

\section*{Acknowledgements}

We thank the Texas A\&M University High Performance Research Computing Center for compute support on the FASTER and Grace clusters, where every experiment in this paper was run.

We also thank the annotators who labeled DamageTriage-Bench for their careful work:
Alex Williams, An Wang, Brendan Ramkissoon, Chris Lee, Connor Nolen, Coy Harris,
Danny Nguyen, Emily Dieu, Eric Fulmer, Eric Robles, Ethan Kishiyama, Hoi-Ling Cheung,
Kaiqi Yang, Sahas Puri, Selina Lee, and Zhuoyi Wang.

\bibliographystyle{plain}
\bibliography{ref}

\end{document}